\newcommand{\CLASSINPUTtoptextmargin}{0.8in}
\newcommand{\CLASSINPUTbottomtextmargin}{0.97in}
\documentclass[conference]{IEEEtran}
\IEEEoverridecommandlockouts

\usepackage{graphicx} % Required for inserting images

\usepackage{mathtools}

\usepackage{dsfont}
\usepackage{cite}
\usepackage{amsmath,amssymb,amsfonts}
\usepackage{nicefrac}

\usepackage{color,soul}

\usepackage[ruled,linesnumbered]{algorithm2e}

\usepackage{algorithmic}

\usepackage{graphicx}
\usepackage{textcomp}
\usepackage{xcolor}

\usepackage{mathtools, stmaryrd}
\usepackage{xparse} 
\usepackage{mathtools, nccmath,amsmath,amssymb,amsfonts,amstext, dsfont, color}
\usepackage{subcaption}
\usepackage{float}

\usepackage{array,multirow}

\newtheorem{theorem}{Theorem}
\newtheorem{corollary}{Corollary}

\newtheorem{remark}{Remark}

\newtheorem{assumption}{Assumption}

\newcommand{\E}[1]{\mathds{E}\left[#1\right]}
\newcommand{\Ex}[2]{\mathds{E}_{#2}\left[#1\right]}

\newcommand{\norm}[1]{\lVert #1 \rVert}
\newcommand{\scalar}[2]{\langle #1 , #2  \rangle}
\newcommand{\hatname}[1]{\hat{\text{#1}}}

\begin{document}

\title{No-Regret Caching with Noisy Request Estimates 
}

\author{
\IEEEauthorblockN{Younes Ben Mazziane, Francescomaria Faticanti, Giovanni Neglia, Sara Alouf}
\IEEEauthorblockA{\textit{Inria, Université Côte d'Azur} \\
Sophia Antipolis, France \\
firstname.lastname@inria.fr}
}
\maketitle

\begin{abstract}
Online learning algorithms have been successfully used to design caching policies with regret guarantees. Existing algorithms assume that the cache knows the exact request sequence, but this may not be feasible in high load and/or memory-constrained scenarios, where the cache may have access only to sampled requests or to approximate requests' counters. In this paper, we propose the Noisy-Follow-the-Perturbed-Leader (NFPL) algorithm, a variant of the classic Follow-the-Perturbed-Leader (FPL) when request estimates are noisy, and we show that the proposed solution
has sublinear regret under specific conditions on the requests estimator. The experimental evaluation compares the proposed solution against classic caching policies and validates the proposed approach under both synthetic and real request traces.

\end{abstract}


\section{Introduction}\label{sec:intro}
Caching techniques are extensively employed in computer systems, serving various purposes such as accelerating CPU performance~\cite{tse1998cpu} and enhancing user experiences in content delivery networks (CDNs)~\cite{bektas2007designing}. The primary objective of a caching system is to carefully choose files for storage in the cache to maximize the proportion of file requests that can be fulfilled locally. This approach effectively minimizes the dependence on remote server retrievals, which can be costly in terms of delay and network traffic.
The presence of caching systems facilitates more efficient data delivery in network traffic and leads to enhanced overall system performance, especially with the widespread adoption of traffic-intensive applications such as virtual and augmented reality~\cite{chatzopoulos2017mobile}, or edge video analytics~\cite{ananthanarayanan2017real}.

Caching policies have been thoroughly investigated under numerous assumptions concerning the statistical regularity of file request processes~\cite{IRM,traverso2013temporal}. However, real-world request sequences tend to deviate from these theoretical models, especially when aggregated over small geographic areas~\cite{leconte2016placing}. This deviation has inspired the exploration of online learning algorithms, beginning with the work of Paschos et al.~\cite{paschos2019learning}, which applied the Online Convex Optimization (OCO) framework~\cite{zinkevich2003online} to caching. These algorithms exhibit robustness to varying request process patterns, as they operate under the assumption that requests may be generated by an adversary.

%With the ongoing increase in network traffic, driven by the widespread adoption of innovative services like virtual and augmented reality applications~\cite{chatzopoulos2017mobile} and edge video analytics~\cite{ananthanarayanan2017real}, caching policies play a vital role in addressing data transportation challenges. By doing so, these policies facilitate more efficient data delivery and lead to enhanced overall system performance.

In this context, the main metric of interest is the \emph{regret}, which is the difference between the cost---e.g., the number of cache misses---incurred by a given online caching algorithm
and the cost of the optimal static cache allocation with hindsight, i.e., with knowledge of the future requests over a fixed time horizon. In this framework, the primary objective is to design \emph{no-regret} algorithms, i.e., online policies whose regret grows sublinearly with the length of the time horizon~\cite{paschos2019learning}.

Several online caching policies have been proposed in the literature, drawing on well-known online algorithms such as Online Gradient Descent (OGD)~\cite{paschos2019learning}, Follow-the-Regularized-Leader (FRL)~\cite{mhaisen2022online} or Follow-the-Perturbed-Leader (FPL)~\cite{mhaisen2022optimistic}. The latter is especially promising, as cache updates can be performed without the need for computationally intensive projection operations over the set of feasible cache states.

% No-regret caching policies have  been studied under the assumption that the cache has perfect knowledge of each request (the only exception we are aware of is~\cite{liu2022learning} discussed below). However, in practical scenarios, the cache may only have access to noisy estimates of the number of requests for each file during a given time window. This can occur due to memory constraints necessitating the use of approximate sketches for counting requests~\cite{einziger2017tinylfu,mazziane2022analyzing} and/or to the high request rate forcing the sampling of the request sequence~\cite{wang2020stcs}.
% % (each request is served but the corresponding counter is not necessarily updated)
% A partial observation regime for requests in online caching is described in~\cite{liu2022learning}, however, the work just considers the case where the caching system is only aware of the requests for already-cached files. This can be seen as a particular case of the most general one of noisy requests considered in our study.

Caching policies, including no-regret ones,  make admission and eviction decisions based on information from the request sequence. This can include factors such as the number of past requests for each file or a list of the most recently requested files. However, when dealing with a vast file catalog and/or a high request rate, a cache might have to depend on noisy information. For instance, limited availability of high-speed memory can necessitate the use of approximate counters based on hash functions~\cite{einziger2017tinylfu,mazziane2022analyzing}. Alternatively, request sampling might be employed to decrease the frequency of counter updates~\cite{li2016full}.

Surprisingly, much of the existing literature on no-regret caching policies overlooks these practical constraints. Typically, these studies operate under the assumption that caches have exact knowledge of the request sequences or of its summaries. A notable exception is the work presented in~\cite{liu2022learning}. However, it exclusively examines the scenario in which the cache is only aware of requests for files it already contains.

In this paper, we bridge this gap by adapting the FPL algorithm---renowned for its computational efficiency and no-regret properties---to manage noisy request estimates.
%In this paper, we address this gap and modify the FPL algorithm, known for its computational efficiency and no-regret properties in caching applications, to handle noisy request estimates. 
Our main contributions are the following:
\begin{itemize}
\item We modify the FPL algorithm to handle noisy request estimates and prove that, under specific conditions on the  estimator, the algorithm maintains sublinear regret. We refer to this extended version as Noisy-Follow-the-Perturbed-Leader (NFPL).
\item We propose two variants of the NFPL algorithm for the caching problem, namely, NFPL-Fix and NFPL-Var, where the requests estimator uses sampling. We prove that NFPL-Fix and NFPL-Var have sublinear regret.   
\item We prove a new regret bound for the classic FPL caching policy that is independent of the catalog size.
\item We show through experimental analysis the advantage of the NFPL algorithm over classical caching policies. We also evaluate the impact of the sampling rate on the performance of NFPL-Fix and NFPL-Var.  
\end{itemize}
The paper is organized as follows. We describe the system assumptions and give  background details in Section~\ref{sec:bacground}. The extension of FPL to deal with noisy requests and its analysis are described in Section~\ref{sec:extending}. Experimental results are presented in Section~\ref{sec:experiments}. Finally, Section~\ref{sec:conclusions} concludes the paper.

\section{System Description and Background}\label{sec:bacground}

\subsection{Caching Problem: Model and Notation}
\label{ss:Caching-problem}
We consider  
a single-cache system in which 
file requests for a catalog $\mathcal{I}$ with $N$ files can either be served locally by a cache with finite capacity $C$ or, in the case of a file miss, by a remote server. 

{\noindent \bf Cache state}. The local cache has a capacity $C \in \{1,\ldots,N\}$ and stores files in their entirety. The cache state at time $t$ is represented by the vector $\mathbf{x}_t = [x_{t,i}]_{i\in \mathcal{N}}$, which indicates the files missing in the cache; that is, $x_{t,i} = 1$ if and only if file $i$ is not stored in the cache at time $t$.
%\footnote{This choice simplifies the following formula like~\eqref{eq:cost}.}
%\footnote{The natural choice would be to indicate with $x_{t,i} = 1$ the presence of item $i$ in the cache, but the adopted one simplifies the following formula like in~\eqref{eq:cost}.}
A feasible cache allocation is then represented by a vector in the set:
\begin{equation}\label{eq:Capped-Simplex}
\mathcal{X} = \left \{ \mathbf{x} \in \{0,1\}^{N} \Bigg| \sum_{i=1}^N x_{i} = N-C \right \}.
\end{equation}

{\noindent \bf Cache updates}. Although caching policies are often assumed to update their state after each request, in high request rate regimes or when cache updates are computationally or communicationally expensive, the cache may update its state after receiving a batch of $B$ requests~\cite{si2023no}.
We study caching policies in this more general setting and consider a time-slotted operation. At each time slot $t=1,\ldots,T$,
$B$ requests are collected from the users and the cache state is updated. The request process is represented as a sequence of vectors $\mathbf{r}_t = (r_{t,i} \in \mathbb{N}: i \in \mathcal{I})$ $\forall t$, where $r_{t,i}$ is the number of requests received for file $i$ in the $t$-th batch.
%at time $t$ in a  batch of size $B$. 
Then, each vector belongs to the set:
%All the requests belong to the set
\begin{equation}\label{eq:requests_set}
%G: why this bar?
\mathcal{B} = \left \{ \mathbf{r}\in  \mathbb{N}^N \Bigg| \;  \sum_{i=1}^N r_{i} = B  \right \}.
% \mathcal{B} =  \{ \mathbf{r}\in  \mathbb{N}^N : \;  \sum_{i=1}^N r_{i} = B  \}.
\end{equation}

{\noindent \bf Cost}. At each time slot $t$, the cache pays a cost equal to the number of misses, i.e., to the number of requests for files not in the cache.
%the number of files that must be retrieved by the remote server. 
The cost can be computed as follows:
\begin{equation}\label{eq:cost}
 \scalar{\mathbf{r}_t}{\mathbf{x}_t}= \sum_{i=1}^N r_{t,i} x_{t,i}\ ,
\end{equation}
where $\scalar{\mathbf{r}}{\mathbf{x}} \triangleq \sum_{i=1}^N r_{i} x_i$ denotes the scalar product of the two vectors $\mathbf{r}$ and $\mathbf{x}$.

For the sake of conciseness, we introduce the following  notation. For any vector $\mathbf{r}$, we denote by $M(\mathbf{r})$ an arbitrary element of $\arg\min_{\mathbf{x}\in \mathcal{X}} \scalar{\mathbf{r}}{\mathbf{x}}$. Furthermore, given a sequence of vectors $(\mathbf{r}_1, \dots, \mathbf{r}_t)$, we represent their aggregate sum as $\mathbf{r}_{1:t}\triangleq \sum_{s=1}^t \mathbf{r}_s$.

\begin{table}
    \centering
    \begin{tabular}{|p{0.1\linewidth}|p{0.6\linewidth}|}
        %\hline
        %Notation & Description \\
        \hline
        $N$ & catalog size \\
        $C$ & cache capacity \\
        $B$ & number of requests in each batch \\
        $\mathcal{X}$ & decision set \\
        $\mathcal{B}$ & set of request vectors \\
        $T$ & time horizon \\
        $\mathbf{r}_t$ & request vector at time step $t$ \\
        $\mathbf{x}_t$ & decision vector at time step $t$ \\
        $\langle \mathbf{r}_t,\mathbf{x}_t \rangle$ & cost at time step $t$ \\
        $\mathbf{r}_{1:t}$ & sum of $\mathbf{r}_s$ for all values of $s$ from $1$ to $t$ \\
        $M(\mathbf{r})$ & value of $\mathbf{x}$ in $\mathcal{X}$ that minimizes $\langle \mathbf{r},\mathbf{x} \rangle$ \\
        $\mathcal{R}_T(\mathcal{A})$ & regret algorithm $\mathcal{A}$ \\
        $\boldsymbol{\gamma}_t$ & noise vector \\
        $\hat{\mathbf{r}}_t$ & noisy request estimates \\
        $\hat{\mathcal{B}}$ & $\hat{\mathbf{r}}_t$ state space \\
        % $\hat{A}$ & supremum of the $\ell_1$-norm of $\hat{\mathbf{r}}$ in $\mathcal{\hat{B}}$  \\
        % $\hat{R}$ & supremum of $\langle \hat {\mathbf{r}},\mathbf{x} \rangle$ for $\mathbf{x}\in \mathcal{X}$ and $\hat {\mathbf{r}}\in \hat{\mathcal{B}}$ \\
        % $D$ & diameter of $\mathcal{X}$ \\
        \hline
    \end{tabular}
    \caption{Table of notation}
    \label{tab:notation}
\end{table}

\subsection{Caching and Online Learning\color{black}}
Caching can be framed as an online learning problem~\cite{hazan2016introduction}, where an agent (the caching system) chooses an action~$\mathbf{x}_t$ from the set~$\mathcal{X}$ at each time slot~$t$ before an adversary reveals a request vector~$\mathbf{r}_t$ from the set $\mathcal{B}$.

The cache state is determined by an online algorithm $\mathcal{A}$ that, at each time slot $t$, computes the cache state $\mathbf{x}_{t+1}$ for the next time slot given the current state $\mathbf{x}_t$ and the sequence of requests up to time~$t$, that is~$\{\mathbf{r_{s}}\}_{s=1}^{t}$. 

The main performance metric used to evaluate an online deterministic algorithm $\mathcal{A}$ choosing action $\mathbf{x}_{t}$ at each time step $t$ is the regret defined as: 
\begin{equation}\label{eq:regret}
    \mathcal{R}_T(\mathcal A) = \sup_{\{\mathbf{r}_1,\ldots,\mathbf{r}_T\}} \left \{\sum_{t=1}^{T} \scalar{\mathbf{r}_t}{\mathbf{x}_t} - \text{OPT}_T\right \},
\end{equation}
where $\text{OPT}_T = \scalar{\mathbf{r}_{1:T}}{M(\mathbf{r}_{1:T})}$ is the cost incurred under the request sequence $\{\mathbf{r}_1,\ldots,\mathbf{r}_T\}$ by the optimal static allocation $\mathbf{x}^* = M(\mathbf{r}_{1:T})$. When the algorithm $\mathcal{A}$ is randomized, one can define the
%similarly to the regret metric, the 
expected regret:
\begin{equation}\label{eq:pseudoregret}
    \mathcal{R}_T(\mathcal A) = \sup_{\{\mathbf{r}_1,\ldots,\mathbf{r}_t\}} \left \{ \E{\sum_{t=1}^{T} \scalar{\mathbf{r}_t}{\mathbf{x}_t}} - \text{OPT}_T\right \}\; ,
\end{equation}
where the expectation is taken over any random choice of the algorithm $\mathcal A$. The expected regret quantifies then the performance gap over a time horizon $T$ between the algorithm $\mathcal{A}$ and the best static cache allocation with hindsight. 

Given the supremum taken over all request sequences in both \eqref{eq:regret} and \eqref{eq:pseudoregret}, it is evident that the regret metrics refrain from making any assumptions regarding the characteristics of the request sequence, such as any inherent statistical regularity.
%Due to the supremum over all request sequences in~\eqref{eq:regret} and in~\eqref{eq:pseudoregret}, regret metrics do not make any assumptions about the nature of the request sequence (e.g., no statistical regularity). 
The request sequence may be thought to have been generated by an adversary seeking to degrade the performance of the caching system. In this setting, one aims for an algorithm with sublinear regret, $\mathcal{R}_T(\mathcal{A}) = o(T)$. These algorithms are commonly known as no-regret algorithms since their time-average cost approaches the optimal static policy's cost as $T$ grows.

Various algorithms, such as Online Gradient Descent (OGD) and Follow-the-Regularized-Leader (FTRL), can attain $\mathcal O(\sqrt{T})$-regret for caching problems~\cite{paschos2019learning,mhaisen2022online}. 
However, their cache update procedures require a computationally expensive projection of a tentative solution back onto the feasible set $\mathcal X$ (e.g., its cost is $O(N^2)$ for OGD~\cite{si2023no}).

In the next section, we present a lightweight caching algorithm with $\mathcal{O}(\sqrt{T})$-regret.

\subsection{Follow-the-Perturbed-Leader (FPL)}
Within the domain of online learning, the Follow-the-Perturbed-Leader (FPL) algorithm is a notable projection-free methodology known to achieve sublinear regret. This algorithm was initially introduced by Vempala et al.~\cite{kalai2005efficient}, and later studied within the caching framework by Bhattacharjee et al.~\cite{bhattacharjee2020fundamental}. 

The FPL algorithm serves as a refined version of the traditional Follow-the-Leader (FTL) algorithm~\cite{littlestone1994weighted}. The latter greedily selects the state that would have minimized the past cumulative cost, i.e., $\mathbf{x}_{t+1}(\text{FTL}) = M(\mathbf{r}_{1:t})$. 

While the FTL algorithm proves optimal when cost functions are sampled from a stationary distribution, it, unfortunately, yields linear regret in adversarial settings~\cite{de2014follow}.

The FPL algorithm improves the performance of FTL by incorporating
a noise vector $\boldsymbol{\gamma}_t$ at each time step $t$.
This vector's components are independent and identically distributed (i.i.d.) random variables, pulled from a distinct distribution (such as the uniform and exponential distributions in~\cite{kalai2005efficient}, and the Gaussian distribution in~\cite{bhattacharjee2020fundamental}). The update process unfolds similarly to FTL:

 \begin{equation}
    \label{e:fpl_update}
     \mathbf{x}_{t}(\text{FPL}) = M(\mathbf{r}_{1:t-1}+\boldsymbol{\gamma}_{t}) .
 \end{equation}

As shown in~\cite{bhattacharjee2020fundamental}, FPL provides optimal regret guarantees for the discrete caching problem. Moreover, the cache update, as specified in equation~\eqref{e:fpl_update}, involves storing the files that correspond to the largest elements of the vector $\mathbf{r}_{1:t-1}+\boldsymbol{\gamma}_{t}$. FPL cache update necessitates then a sorting operation.
Notably, its computational complexity of $\mathcal O(N \log N)$ is less taxing than the projection step required by either the FRL or OGD algorithms, as highlighted in~\cite{bhattacharjee2020fundamental}.
%whose $\mathcal O(N \log N)$ is less demanding than 
%. This operation is computationally less demanding ($\mathcal $)than 
%the projection step required by the FRL or OGD algorithms~\cite{bhattacharjee2020fundamental}.

\section{{Extending FPL}}\label{sec:extending}

\begin{algorithm}[t]
\KwIn{Set of decisions $\mathcal{X}$; $T$; $\eta$}
\KwOut{Sequence of decisions: $\{\mathbf{x}_t\}_{1}^{T}$}

$\hat{\textbf{costs}} \gets  0$ ,  \\ 
\For{round $t=1,2,...,T$}{
$\boldsymbol{\gamma}_t \sim \textbf{Unif}\left(\left[0, \eta\right]^{N}, \mathbb{I}_{N\times N}\right)$ \\
$\mathbf{x}_t \gets M(\hat{\textbf{costs}} +  \boldsymbol{\gamma}_t)$ \\
Pay $\langle \mathbf{r}_t,  \mathbf{x}_t \rangle$ \\ \label{line:noisy-request}
Observe $\hat{\mathbf{r}}_{t}$ \\ 
$\hat{\textbf{costs}} \gets \hat{\textbf{costs}} + \mathbf{\hat{r}}_t$
}
\caption{Noisy-Follow-the-Perturbed-Leader with Uniform Noise (NFPL)}
\label{alg:NFPL}
\end{algorithm}
The traditional FPL algorithm needs to track the request count for each file in the catalog.
%, a difficult task in environments where the arrival rate of requests is very high
As we discussed in the introduction, in scenarios with a large catalog and/or high request rate, the cache may only have access to noisy estimates.
%Applying the FPL algorithm to the caching problem can be practically challenging. In fact, FPL monitors the request count for each file in the catalog, a difficult task in environments where the arrival rate of requests is very high. Sampling the request process is a common practice in high-rate request regimes at the expense of noisy request estimates~\cite{li2016full}. 
For this reason, we introduce the Noisy-Follow-the-Perturbed-Leader (NFPL), a lightweight variant of FPL that employs noisy request estimates instead of exact request counts. In Section \ref{ss:NFPL-Details}, we present the NFPL algorithm in detail along with its regret analysis. Subsequently, in Section~\ref{ss:NFPL-Caching}, we study NFPL when noisy request estimates stem from sampling the request process as in~\cite{li2016full}.

% The FPL algorithm applied to the caching problem involves monitoring the request count for each file in the catalog. In a partial observation regime, for example, \cite{liu2022learning}, the request count for each file is noisy. This also occurs in environments where the arrival rate of requests is very high forcing the sampling of the requests~\cite{netflow2005}. For this reason, we introduce the Noisy-Follow-the-Perturbed-Leader (NFPL), a lightweight variant of FPL that employs noisy request estimates instead of exact request counts. In Section \ref{ss:NFPL-Details}, we present the NFPL algorithm in detail along with its regret analysis, and in Section~\ref{ss:NFPL-Caching}, we apply the NFPL algorithm for the caching problem where the noisy request estimates stem from sampling the request process. 

\subsection{Noisy-Follow-the-Perturbed-Leader (NFPL)}
\label{ss:NFPL-Details}
The NFPL algorithm is described in Algorithm~\ref{alg:NFPL}. NFPL follows in the footsteps of FPL with uniform noise but observes the estimated requests
$\mathbf{\hat r}_t$ instead of the real requests
$\mathbf{r}_t$. In particular, at each time slot $t$, the algorithm generates $\boldsymbol{\gamma}_t$ from a multivariate uniform distribution with uncorrelated components, constrained within the range $[0, \eta]^N$,
and it updates the decision vector $\mathbf{x}_t$ with the minimizer of $\scalar{\mathbf{x}}{\mathbf{\hat r}_{1:t-1} +  \boldsymbol{\gamma}_t}$ over $\mathbf{x} \in \mathcal{X}$. The cost paid at time slot $t$ is equal to $\langle \mathbf{r}_t, x_t\rangle$. The total cost of the NFPL algorithm is 

 \begin{align}
        \text{NFPL}_{T} = \sum_{t=1}^{T} \scalar{\mathbf{r}_t}{ M(\mathbf{\hat r}_{1:t-1} + \boldsymbol{\gamma}_t)} .
\end{align}

We remark that Algorithm \ref{alg:NFPL} is not just confined to the caching scenario discussed in Section~\ref{sec:bacground}. Indeed, it is also applicable to any situation where the agent incurs costs represented by the equation $\scalar{\mathbf{r}_t}{\mathbf{x}_t}$.

\begin{assumption} \label{assum:NFPL-sublinear}
    We assume that $\mathbf{\hat r}_t$ is an unbiased estimator of~$\mathbf{r}_t$, which implies that the expected value of~$\mathbf{\hat r}_t$ equals $\mathbf{r}_t$, i.e.,~$\E{\mathbf{\hat r}_t}=\mathbf{r}_t$. 
\end{assumption}

\begin{assumption}\label{assum:Constants-NFPL}
    Let $\mathcal{\hat B}$ be the state space of $\mathbf{\hat r}_t$. We assume the existence of the following constants: 
        \begin{align}
                 &\hat A = \sup_{\mathbf{\hat r} \in \mathcal{\hat B}}\norm{\mathbf{\hat r} }_1,\; \hat R = \sup_{\mathbf{x}\in \mathcal{X}, \mathbf{r} \in \mathcal{\hat{B}}} \scalar{\mathbf{r}}{\mathbf{x}}, \\
                &D= \sup_{\mathbf{x},\mathbf{y} \in \mathcal{X}} \norm{\mathbf{x}-\mathbf{y}}_{1}.
        \end{align}
    
    % $\hat A = \sup_{\mathbf{\hat r} \in \mathcal{\hat B}} \norm{\mathbf{\hat r} }_1$,~$\hat R = \sup_{\mathbf{x}\in \mathcal{X}, \mathbf{r} \in \mathcal{\hat{B}}} \scalar{\mathbf{r}}{\mathbf{x}}$ and~$D= \sup_{\mathbf{x},\mathbf{y} \in \mathcal{X}} \norm{\mathbf{x}-\mathbf{y}}_{1}$ exist.
\end{assumption}

\begin{theorem}[Regret bound NFPL]\label{th:Regret-NFPL}
Under Assumptions \ref{assum:NFPL-sublinear} and \ref{assum:Constants-NFPL}, the NFPL algorithm with $\eta = \sqrt{\hat R\cdot \hat A \cdot T/D}$ enjoys sublinear regret:
        \begin{align} \label{e:NFPL-regret-bound}
                 \mathcal{R}_{T}(\text{NFPL}) \leq 2 \sqrt{\hat R\cdot \hat A \cdot D\cdot T} . 
        \end{align}
\end{theorem}

\begin{IEEEproof}
%Let $\{\mathbf{r}_t\}_{1}^{T}$ be a sequence of requests vectors, $\{\mathbf{\hat r}_t\}_{1}^{T}$ a realization from $\{\mathbf{r}_t\}_{1}^{T}$'s unbiased estimator and $\{\boldsymbol{\gamma}_t\}_{1}^{T}$ realizations from a multivariate uniform distribution with uncorrelated components, constrained within the range $[0, \eta]^N$. 
   % The cost of NFPL given $\{\mathbf{r}_t, \mathbf{\hat r}_t , \boldsymbol{\gamma}_t\}_{1}^{T}$ is expressed as: 
   %  \begin{align}
   %      \text{NFPL}_{T} = \sum_{t=1}^{T} \scalar{\mathbf{r}_t}{ M(\mathbf{\hat r}_{1:t-1} + \boldsymbol{\gamma}_t)} .
   %  \end{align}
It is convenient to define the following two auxiliary quantities
\begin{align}
  &\hatname{NFPL}_{T} = \sum_{t=1}^{T} \scalar{\mathbf{\hat r}_t} {M(\mathbf{\hat r}_{1:t-1} + \boldsymbol{\gamma}_t)},\\
  &\hatname{OPT}_{T} = \scalar{\mathbf{\hat r}_{1:T}}{M(\mathbf{\hat r}_{1:T})}. 
 \end{align}
We compute the expectation---over $\{\mathbf{\hat r}_t , \boldsymbol{\gamma}_t\}_{1}^{T}$---of the difference between the total cost of NFPL and $\text{OPT}_{T}= \scalar{\mathbf{r}_{1:T}}{M(\mathbf{r}_{1:T})}$ as follows
\begin{align} \nonumber
 &\E{\text{NFPL}_{T} - \text{OPT}_{T}} \\  \nonumber
&= \E{\text{NFPL}_{T} - \hatname{NFPL}_{T}}  + \E{\hatname{NFPL}_{T} - \hatname{OPT}_{T}}\\ \label{e:proof-NFPL4}
&\quad + \E{ \hatname{OPT}_{T} -\text{OPT}_{T}} .
\end{align}
We have: 
\begin{align} \label{e:proof-NFPL1}
    &\E{\text{NFPL}_{T} - \hatname{NFPL}_{T}} =0 ,\\ \label{e:proof-NFPL2}
    &\E{\hatname{OPT}_{T}-\text{OPT}_{T}} \leq 0 ,\\ 
    \label{e:proof-NFPL3}
    &\E{\hatname{NFPL}_{T} - \hatname{OPT}_{T}} \leq 2\sqrt{\hat R\cdot \hat A\cdot D\cdot T } .
\end{align}
The random vectors $\mathbf{\hat {r}}_t$ and $M(\mathbf{\hat {r}}_{1:t-1} +\boldsymbol{\gamma}_t)$ are independent, hence $\E{\scalar{\mathbf{\hat r}_t}{M(\mathbf{\hat r}_{1:t-1} + \boldsymbol{\gamma}_t)} } = \scalar{\mathbf{r}_t}{\E{M(\mathbf{\hat r}_{1:t-1}+\boldsymbol{\gamma}_t)}}$ and by linearity of the expectation we deduce \eqref{e:proof-NFPL1}. 
We have that $\hatname{OPT}_T\leq \scalar{\mathbf{\hat r}_{1:T}}{M(\mathbf{r}_{1:T})} $, we get then \eqref{e:proof-NFPL2}.  

The quantity $\hatname{NFPL}_{T} - \hatname{OPT}_{T}$ can be seen as the difference between the cost of an FPL algorithm with uniform noise, that observes costs $\{\mathbf{\hat{r}}_t\}_{1}^{T}$, minus the cost incurred by the optimal static allocation $\mathbf{x}^{*} = M(\mathbf{\hat r}_{1:T})$. Therefore, applying \cite[Theorem 1.1 a)]{kalai2005efficient} with $\epsilon = 1/\eta$ such that $\eta = \sqrt{\hat R\cdot \hat A \cdot T/D}$, we get
\begin{equation*}
\Ex{\hat{\text{NFPL}}_{T} - \hat{\text{OPT}}_{T}}{\{\boldsymbol{\gamma}_t\}_{1}^{T}} \leq 2\sqrt{\hat R \cdot \hat A \cdot D \cdot T} 
\end{equation*}
for any $\{\mathbf{\hat r}_t\}_{1}^{T}$, and by taking the expectation over the randomness of $\{\mathbf{\hat r}_t\}_{1}^{T}$ in both sides of the last inequality, we find \eqref{e:proof-NFPL3}. Plugging \eqref{e:proof-NFPL1}, \eqref{e:proof-NFPL2} and \eqref{e:proof-NFPL3} in \eqref{e:proof-NFPL4}, we deduce that $\E{\text{NFPL}_{T} - \text{OPT}_{T}}\leq 2\sqrt{\hat R\cdot \hat A\cdot D\cdot T}$ for every $\{\mathbf{r}_t\}_{1}^{T}$,  concluding the proof. 

\end{IEEEproof}

\begin{remark}
NFPL regret bound in Theorem 1 can be written as $\alpha \cdot \beta$ where $\alpha = \hat R \cdot \hat A / (R \cdot A) $, $\beta = 2\sqrt{R\cdot A \cdot D\cdot T}$, $R=\sup_{\mathbf{x}\in \mathcal{X}, \mathbf{r} \in \mathcal{\bar B}} \scalar{\mathbf{r}}{\mathbf{x}} $ and $A = \sup_{\mathbf{r}\in \mathcal{\bar B}} \norm{\mathbf{r}}_1$. Observe that $\beta$ is FPL's classical regret bound
when the algorithm knows the exact costs~\cite[Theorem 1.1 a)]{kalai2005efficient}. 
It is easy to verify that $\alpha$ is greater than or equal to 1 and can then be interpreted as the performance loss the algorithm incurs due to the noisy costs. 
%representing a constant that captures the effect of the utilization of noisy requests in the regret bound.

\end{remark}

% We now study three specific cases of the unbiased estimator in the caching problem setting: i) the case where a fixed number of requests is sampled within each batch, ii) the case where each request is sampled with a given probability, iii) $\mathbf{\hat r}_t = \mathbf{r}_t$, i.e., the estimator is deterministic and always correct. 

\begin{figure*}[t]
\begin{subfigure}{0.32\linewidth}
  \centering
  \includegraphics[width=0.99\linewidth, keepaspectratio]{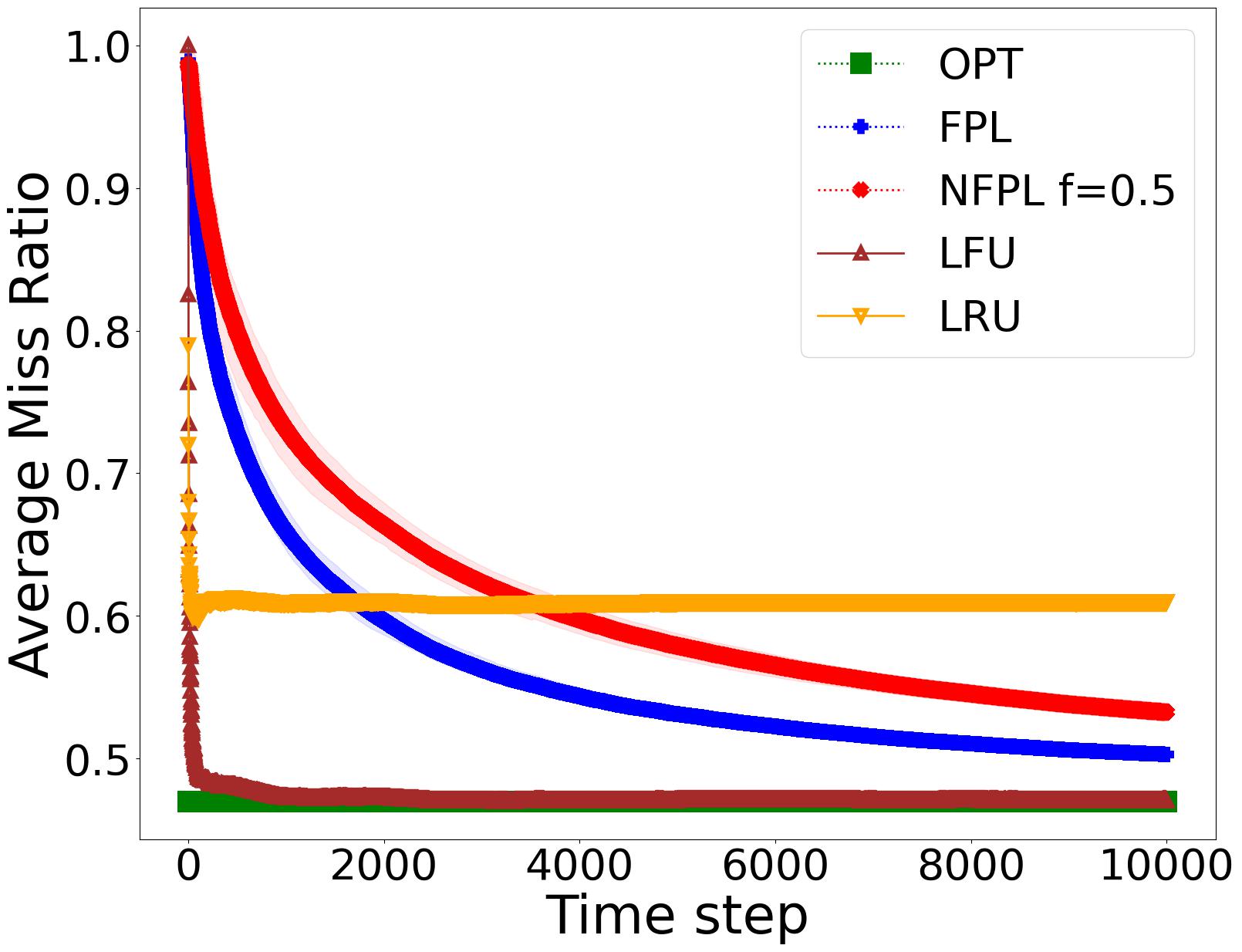}
  \caption{Zipf}
  \label{fig:hit-rate-d2}
\end{subfigure}
\hfill
    \begin{subfigure}{0.32\linewidth}
  \centering
  \includegraphics[width=0.99\linewidth]{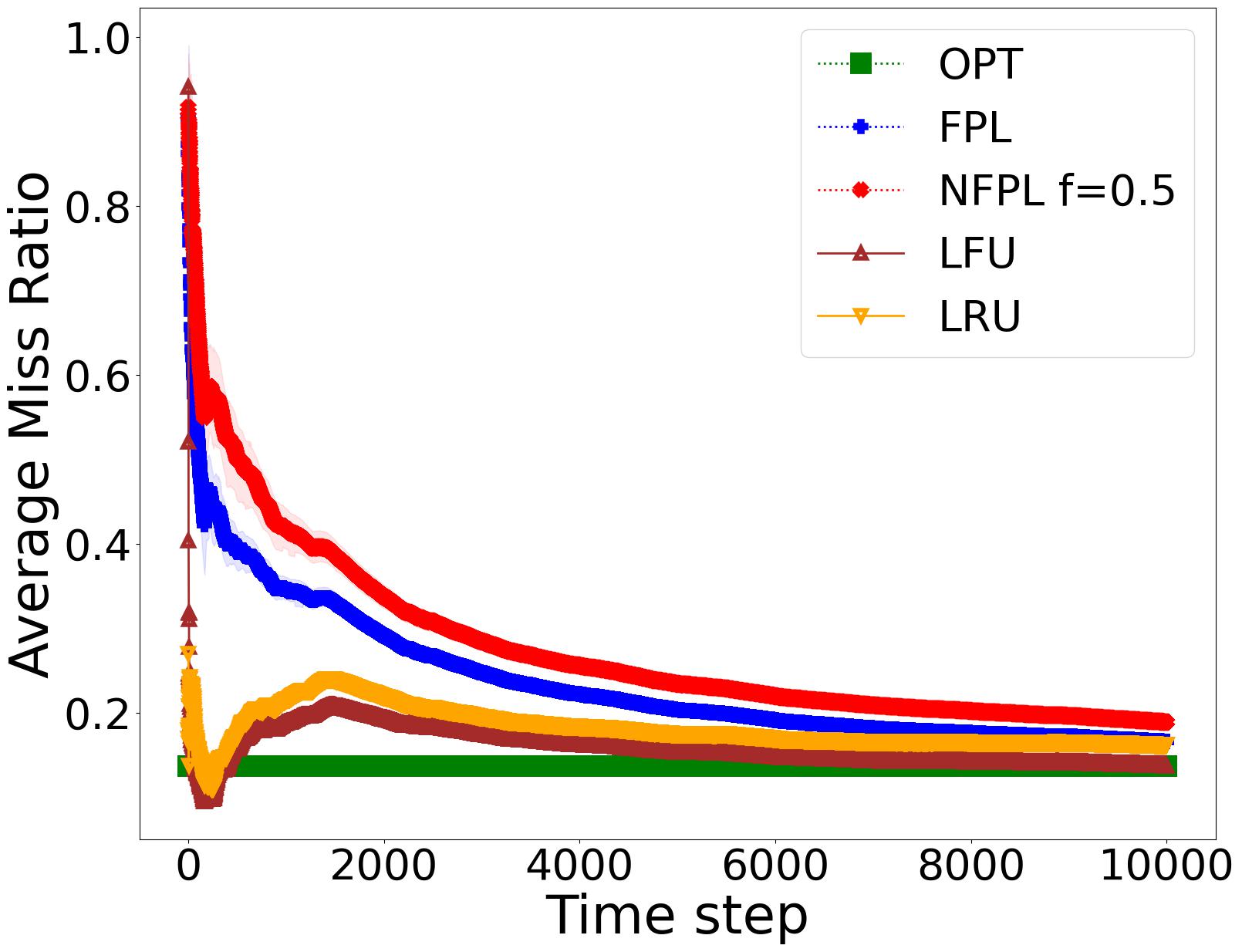}
\caption{Akamai}
\label{fig:NACRRSmall}
\end{subfigure}
\hfill
 \begin{subfigure}{0.32\linewidth}
      \centering  
    \includegraphics[width=0.99\linewidth,keepaspectratio]{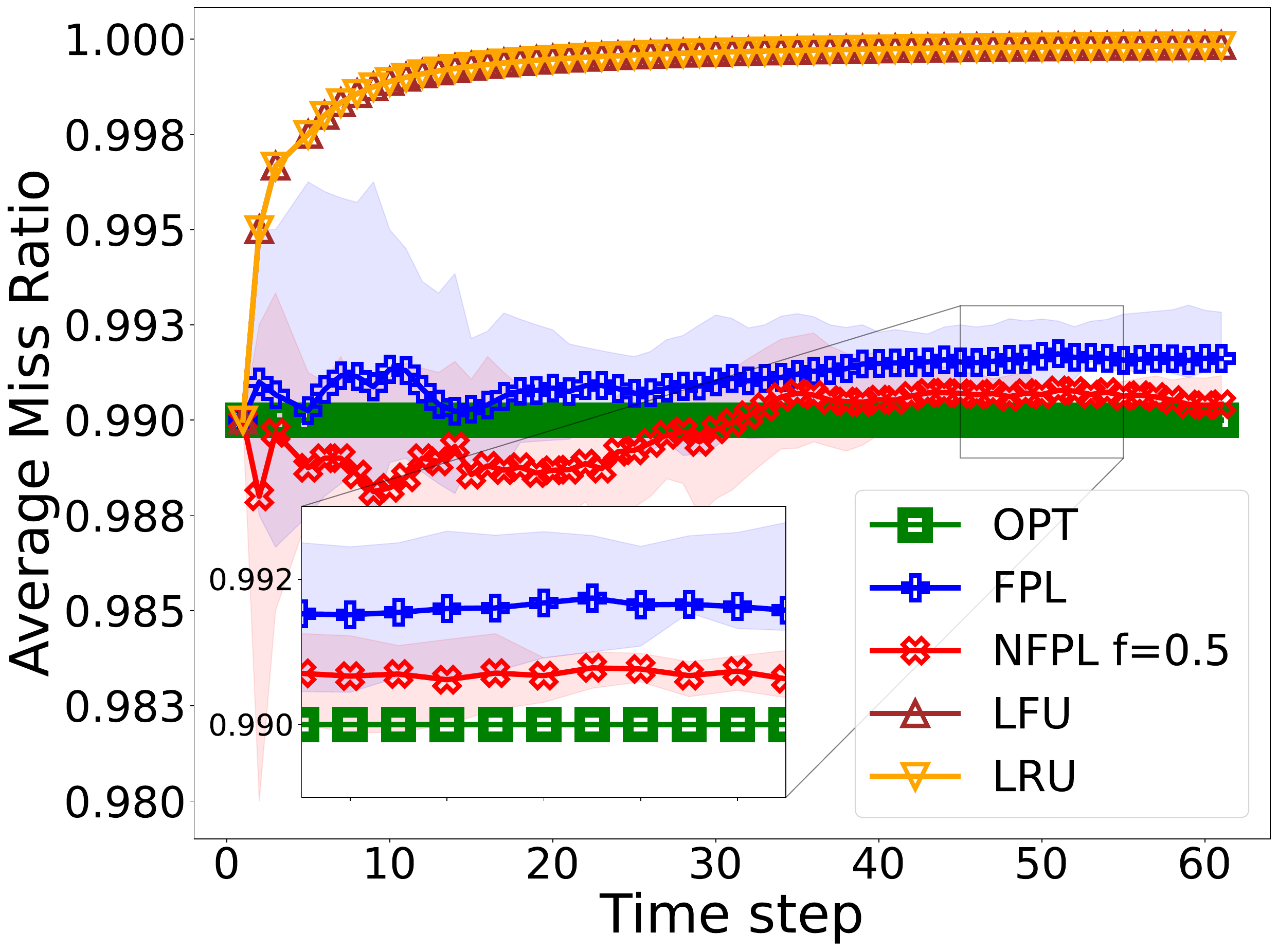}
    \caption{Round-robin}
    \label{fig:Average-Miss-Ratio-Round-Robin}
\end{subfigure}
    \caption{Average miss ratio, $C=100$.}
    \label{fig:hit-rates} 
    \vspace{-0.1in}
\end{figure*}

\subsection{NFPL for Caching}
\label{ss:NFPL-Caching}

 We apply NFPL to the caching problem (section \ref{ss:Caching-problem}), deriving $\mathbf{\hat{r}}_t$ from sampled requests. Two methods are explored: NFPL-Fix, sampling a fixed number of requests within each batch, and NFPL-Var, independently sampling each request within the batch with a fixed probability.

{\noindent \textbf{NFPL-Fix}}. The caching system samples $b\geq 1$ requests uniformly at random from a batch of $B$ requests at each time slot. Let $\mathbf{\hat d}_t$ be the number of requests for each file in the sampled batch at time step $t$. NFPL-Fix is Algorithm \ref{alg:NFPL} with noisy request estimates $\mathbf{\hat r}_t$ given by  
\begin{equation}\label{e:fixed-batch}
        \mathbf{\hat{r}}_t  =  \frac{B}{b} \cdot  \mathbf{\hat{d}}_t. \; \
\end{equation}

\begin{corollary}[Regret bound NFPL-Fix]\label{co:NFPL-caching-fixed-subBatch}
      NFPL-Fix with  $\quad \eta = B \sqrt{2T}/ 2C$ has sublinear~regret:
        \begin{align}
                 \mathcal{R}_{T}(\text{NFPL-Fix}) \leq 2\sqrt{2} \cdot B \sqrt{C\cdot T} . 
        \end{align}
\end{corollary}
\begin{IEEEproof}
Observe that with \eqref{e:fixed-batch}, we have that $\E{\mathbf{\hat r}_t} = \mathbf{r}_t$. Since $\norm{\hat{\mathbf d}_t}_1 =b$, then $\hat A= B$ and $\hat R \leq B$. We have that $D\leq 2C$, hence by applying Theorem \ref{th:Regret-NFPL} in this setting, the regret bound readily follows concluding the proof.
\end{IEEEproof}

% \begin{remark}
% Corollary \ref{co:NFPL-caching-fixed-subBatch} provides the same regret guarantees for all integer values of $b$ larger or equal to $1$. However, we show experimentally in Section~\ref{sec:experiments} that using larger values of $b$ can improve the performance of NFPL-Fix. 
% \end{remark}

{\noindent \textbf{NFPL-Var}}. The caching system samples each request within the batch of requests with a probability $f>0$ at each time slot. Let $\mathbf{\hat s}_t$ be the number of requests for each file in the sampled batch at time step $t$. NFPL-Var is Algorithm \ref{alg:NFPL} with noisy request estimates $\mathbf{\hat r}_t$ expressed as
\begin{equation}\label{e:estimator-rt-f}
    \mathbf{\hat r}_t = \frac{1}{f} \cdot \mathbf{\hat s}_t.  
\end{equation}

\begin{corollary}[Regret bound NFPL-Var]\label{co:NFPL-Var}
       NFPL-Var with $\quad \eta =  B \sqrt{2T}/ \left(2f \cdot C\right)$ has sublinear regret: 
        \begin{align}
                 \mathcal{R}_{T}(\text{NFPL-Var}) \leq 2\sqrt{2} \cdot \frac{B}{f} \cdot \sqrt{C\cdot T}. 
        \end{align}
\end{corollary}
\begin{IEEEproof}
Observe that with \eqref{e:estimator-rt-f}, we have that $\E{\mathbf{\hat r}_t} = \mathbf{r}_t$. Moreover, we observe that the maximum of $\norm{\mathbf{\hat r}_t}_1$ is attained when the sub-batch includes all the requests from the batch, i.e., $\norm{\mathbf{\hat s}_t}_1= B$. It follows that $\hat A = B/f$ and $\hat R \leq B/f$. We have that $D\leq 2C$, hence by applying Theorem \ref{th:Regret-NFPL} in this setting, the regret bound readily follows concluding the proof.

\end{IEEEproof}

% \begin{remark} 
% NFPL-Var maintains sublinear regret as long as the sampling probability $f$ verifies $f=\Omega(1/\sqrt{T})$ (see Corollary \ref{co:NFPL-Var}). However, smaller values of $f$ require a larger set of observed requests to guarantee a target average regret $\epsilon$ ($\mathcal{R}_{T}/T\leq \epsilon)$.

% \end{remark}

For $b=B$ and $f=1$, request counts are exact, i.e., $\mathbf{\hat r}_t = \mathbf{r}_t$, and both NFPL-Fix and NFPL-Var coincide with the classic FPL.
%NFPL-Fix with $b=B$ or NFPL-Var with $f=1$ reduces to the case where the file requests counts are exact, i.e., $\mathbf{\hat r}_t = \mathbf{r}_t$ and the algorithm . 
Using Corollary \ref{co:NFPL-caching-fixed-subBatch} or Corollary \ref{co:NFPL-Var} we deduce the following corollary. 

\begin{corollary} [Regret bound FPL caching] \label{co:FPL-uniform-noise}
%Algorithm \ref{alg:NFPL} applied for caching, with $\mathbf{\hat r}_t = \mathbf{r}_t$ and 
FPL with $\eta =  B \sqrt{2T} / 2C$ has sublinear regret:
\begin{align}
        \label{e:bound_exact_knowledge}
           \mathcal{R}_{T}(\text{FPL})
& \leq 2\sqrt{2}\cdot B \sqrt{C \cdot T}.
\end{align}
\end{corollary}

%In~\cite{bhattacharjee2020fundamental}, Bhattacharjee et al. proposed an FPL-based algorithm with Gaussian noise for the caching problem in the specific case where $B=1$, and derived a bound on the regret. The result is copied next for completeness. 
The authors of~\cite{bhattacharjee2020fundamental} proved regret guarantees for FPL applied to caching under perfect knowledge of the requests when $B=1$. We report the result here for completeness. 
\begin{theorem}\cite[Thm. 3]{bhattacharjee2020fundamental}
\label{th:FPL-gaussian-noise}
FPL applied to the caching problem with $B=1$, learning rate $\eta=\frac{1}{4\pi (\ln{N})^{1/4}} \sqrt{\frac{T}{C}}$, and noise vectors $\{\boldsymbol{\gamma}_t\}_{1}^{T}$, where $\boldsymbol{\gamma}_t/\eta$ is drawn from a standard multivariate normal distribution, has sublinear regret. More specifically
  \begin{align}
        \label{e:bound_bhattacharjee}
      \mathcal{R}_{T}(\text{FPL})  \leq 1.51 \cdot (\ln N)^{1/4} \cdot \sqrt{C\cdot T}.
  \end{align}
\end{theorem}

The comparison of Corollary~\ref{co:FPL-uniform-noise} and Theorem 2 shows that our 
analysis is also of interest when requests are exactly known. First, our bound~\eqref{e:bound_exact_knowledge}  is also valid when the requests are batched, which is of practical interest since updating the cache at each request might be computationally impractical. Second, our bound does not depend on the catalog size, as~\eqref{e:bound_bhattacharjee} does, and in particular it will not diverge as $N$ goes to infinity.

% \section{\textcolor{blue}{Analysis of NFPL}} 
% \input{analysis}

\section{Numerical Evaluation}\label{sec:experiments}

\begin{figure*}[t]
\centering
%\captionsetup[subfigure]{aboveskip=-1pt,belowskip=-1pt}
\begin{subfigure}{0.32\linewidth}
    \centering
    \includegraphics[width=0.95\linewidth, keepaspectratio]{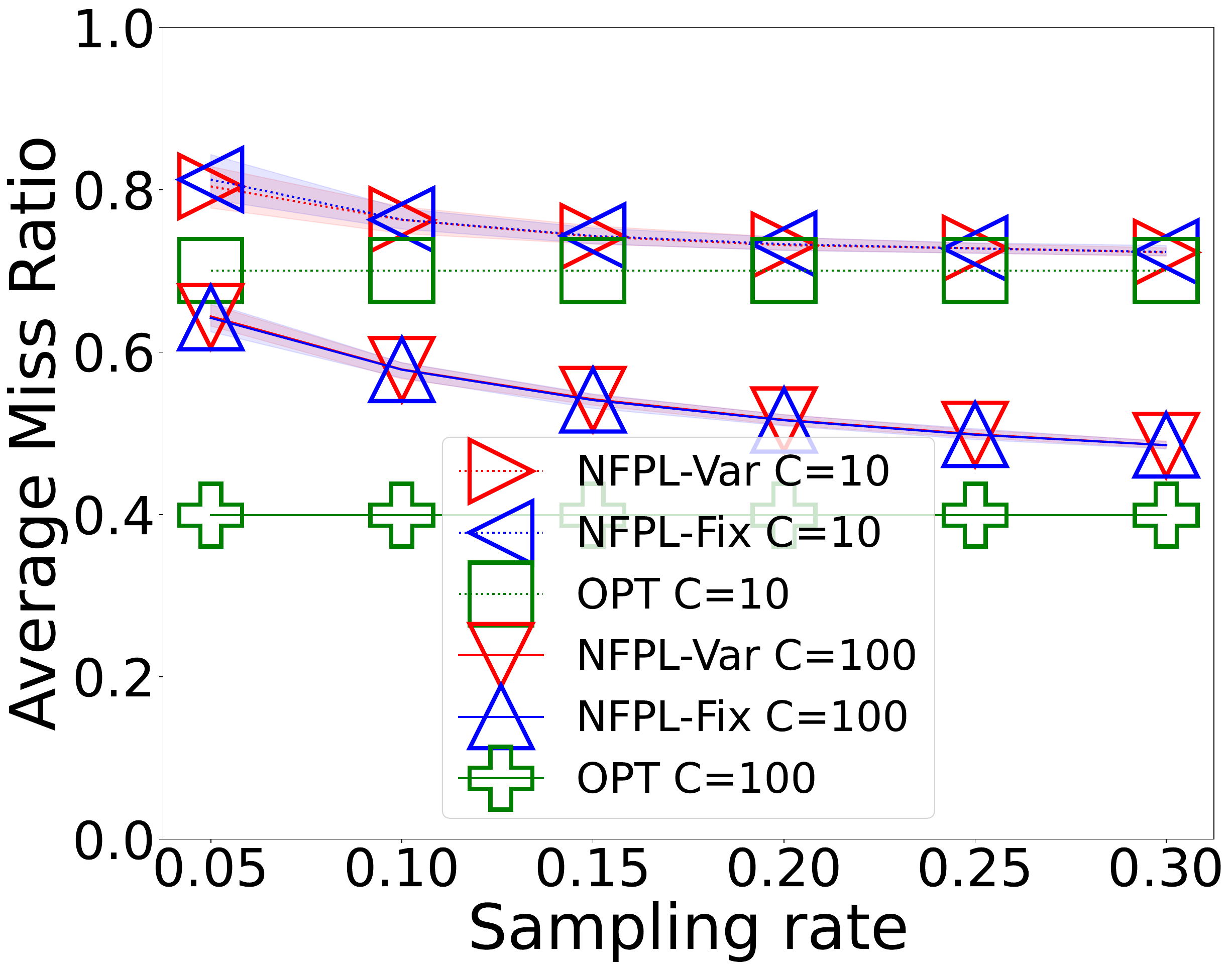}
\caption{Zipf}
\label{fig:Zipf-NFPL-FIX-VAR}
\end{subfigure}
\hfill
\begin{subfigure}{0.32\linewidth}
  \centering
  \includegraphics[width=0.95\linewidth,keepaspectratio]{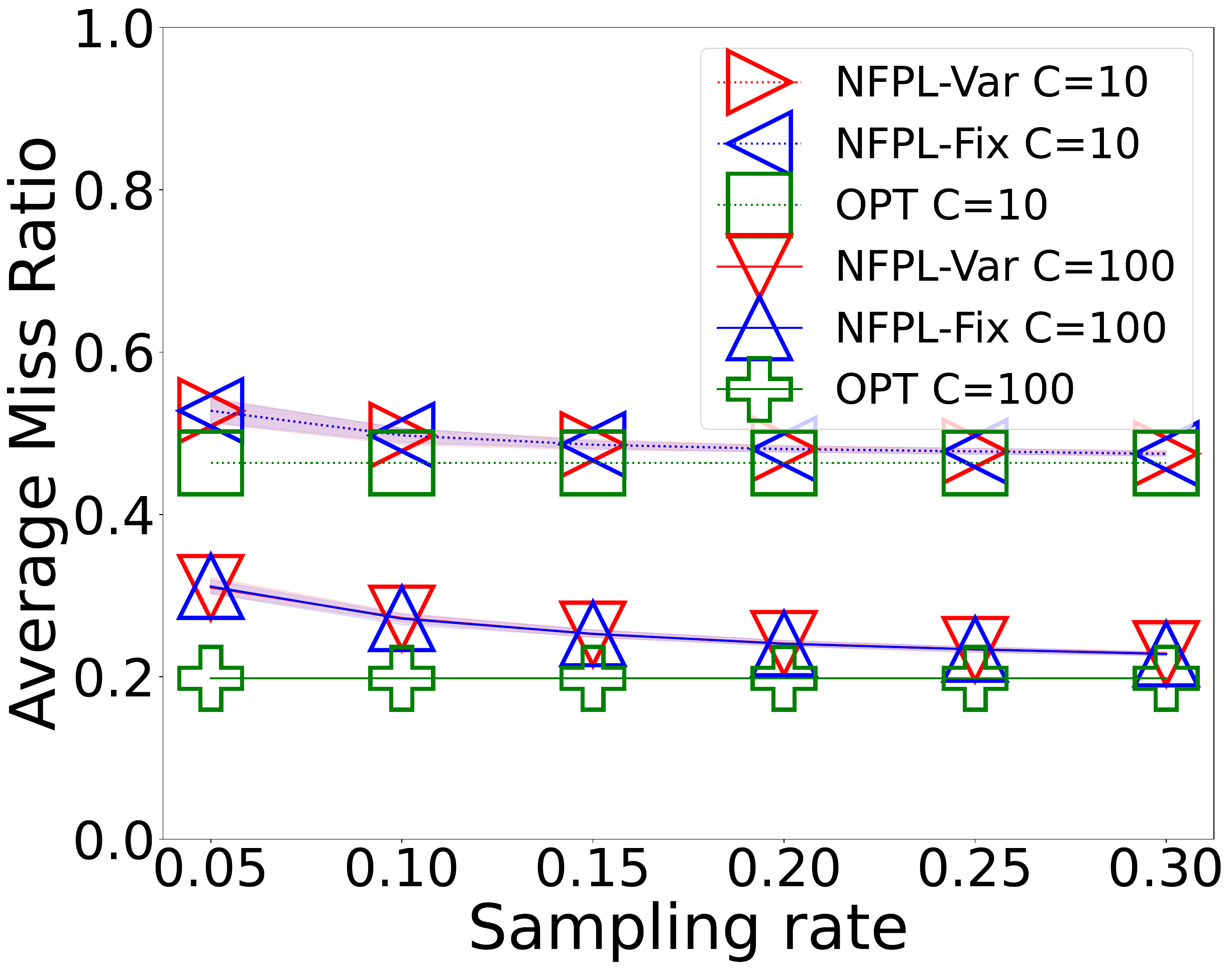}
\caption{Akamai}
\label{fig:Akamai-NFPL-FIX-VAR}
\end{subfigure}
\hfill
\begin{subfigure}{0.32\linewidth}
  \centering
  \includegraphics[width=0.95\linewidth,keepaspectratio]{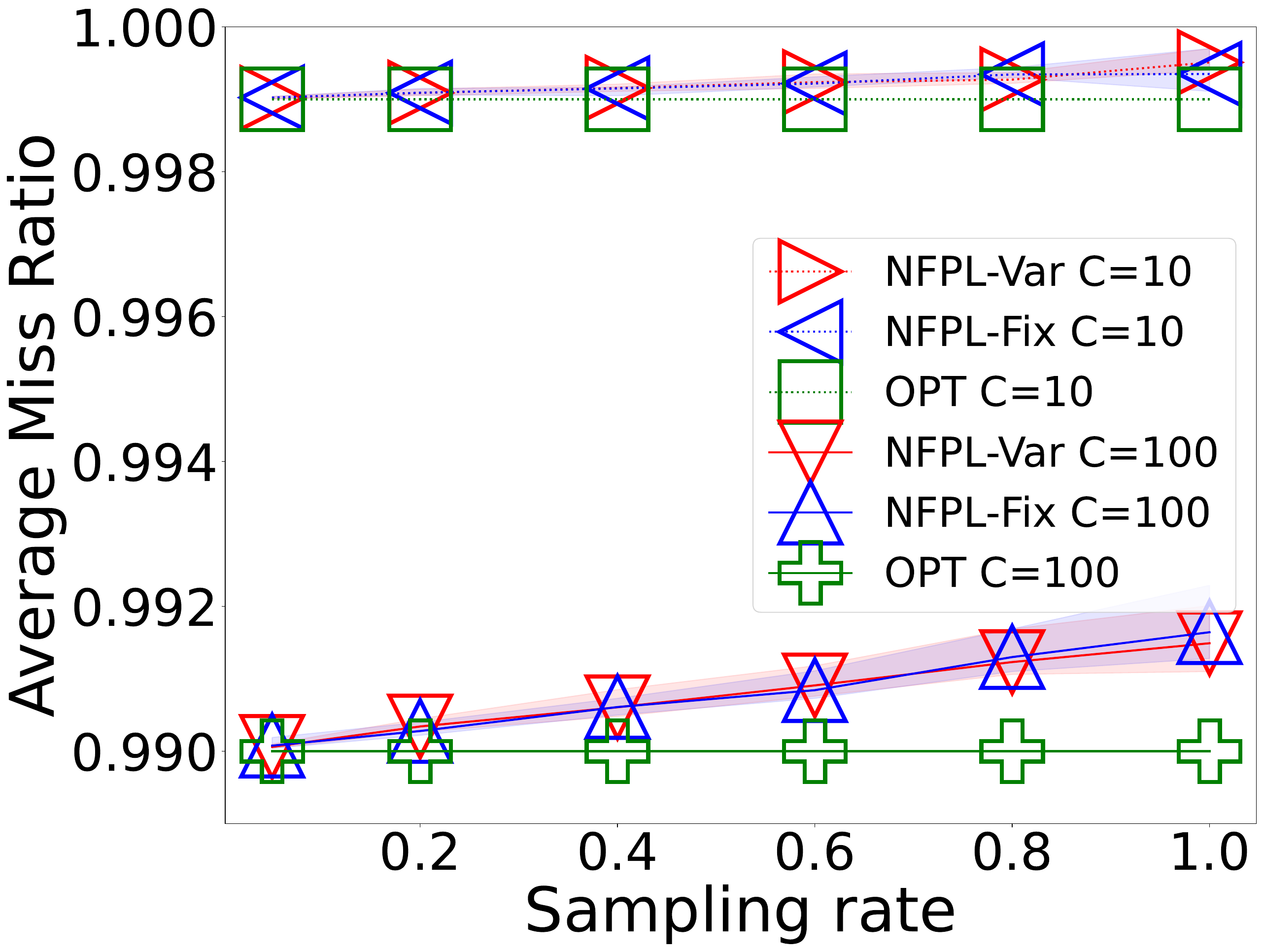}
\caption{Round-robin}
\label{fig:Round-robin-NFPL-FIX-VAR}
\end{subfigure}
\caption{Average miss ratio vs. sampling probability.}
\label{fig:NFPL-FIX-VAR}
\vspace{-0.1in}
\end{figure*}

We conducted simulations of NFPL-Fix and NFPL-Var and other existing policies, using both synthetic and real-world traces. Details about the traces are presented in Section~\ref{ss:Datasets}, while Section~\ref{ss:Caching-policies} discusses the caching baselines. We evaluate the effectiveness of our proposed algorithms, NFPL-Fix and NFPL-Var, from two perspectives. First, in Section~\ref{ss:NFPL-VS-ClassicalPolicies}, we compare the NFPL family of algorithms to traditional caching algorithms. Second, we compare NFPL-Fix and NFPL-Var and show the effect of sampling on their performance in Section~\ref{ss:NFPL-Fix-Vs-NFPL-Var}.

\subsection{Traces}
\label{ss:Datasets}

{\noindent  \bf{Zipf trace.}} We generate a total of $5\times 10^{6}$ requests from a catalog of $N=10^{4}$ files following an i.i.d. Zipf distribution with exponent $\alpha=1$. The Zipf distribution is a popular model for the request process in caching~\cite{breslau1999web}.

{ \noindent \bf{Akamai trace.}} 
The request trace, sourced from Akamai CDN as documented in \cite{neglia2017access}, encompasses several days of file requests, amounting to a total of $2\times 10^{7}$ requests for a catalog comprising $N=10^{3}$ files.

%It is a request trace collected in Akamai CDN~\cite{neglia2017access}, capturing a few days worth of file requests totaling $2\times 10^{7}$ for a catalog of $N=10^{3}$ files.

{\noindent  \bf{Round-robin trace.}} 
We generate a total of $10^{6}$ file requests from a catalog comprising $N=10^4$ files in a round-robin fashion. The round-robin trace is commonly considered as an adversarial trace~\cite{bhattacharjee2020fundamental}.

\subsection{Caching policies}
\label{ss:Caching-policies}

We compare our methods (NFPL-Fix and NFPL-Var) with the optimal static cache allocation with hindsight (OPT), FPL with perfect knowledge of the requests (equivalent to NFPL-Var with $f=1$), as well as two classic caching policies: Least-Frequently-Used (LFU) and Least-Recently-Used (LRU). Upon a miss, LFU and LRU evict from the cache the least popular file and the least recently requested file, respectively.
FPL and NFPL policies are configured with $T$ equal to the number of batches in the corresponding trace.

All the aforementioned caching policies are evaluated with the \textit{average miss ratio} computed as follows
\begin{align}
\frac{1}{Bt}\sum_{\tau=1}^t \scalar{\mathbf{r}_{\tau}}{\mathbf{x}_{\tau}}. 
\end{align}
For NFPL-Fix and NFPL-Var, the average miss ratio is averaged over $M=50$ runs, considering different noisy request estimates $\{\mathbf{\hat r}_t\}_{1}^{T}$ and noise vectors $\{\boldsymbol{\gamma}_t\}_{1}^{T}$. To account for the variability across the runs, we report the first and ninth deciles of the average miss ratio. In all experiments, the batch size $B$ is set to~$200$.

\subsection{NFPL vs. classical policies}
\label{ss:NFPL-VS-ClassicalPolicies}
 
We simulate NFPL-Var, with sampling probability $f=0.5$, FPL, LRU, LFU, and OPT over all the presented traces. In Figure \ref{fig:hit-rates}, we show the average miss ratio at each time step $t$.

In the Zipf trace, files popularity does not change over time and LFU rapidly discerns the most popular files and subsequently converges to OPT. However, due to the noise $\gamma_t$, FPL requires a longer duration to accurately determine the files to be stored. NFPL, on the other hand, grapples with two sources of noise: the inherent noise $\gamma_t$ and the additional noise due to sampling. As a result, NFPL takes even longer to adjust. Nevertheless, both FPL and NFPL outperform LRU, whose missing ratio fails to converge to OPT.

In the Akamai trace, it is plausible to anticipate fluctuations in popularity over time, and requests' temporal correlations. Such patterns can be advantageous for LRU. In fact, LRU now performs almost on par with LFU. 
Notably, both FPL and NFPL appear to be converging to the performance of OPT.

%In the Akamai and Zipf traces, LFU outperforms all the other caching policies. Nevertheless, NFPL-Var and FPL have comparable performances with LFU as time increases. We stress that NFPL-Var, despite the noisy requests, still competes with OPT and even outperforms LRU in the Zipf trace. 

Under the round-robin trace, optimality can be achieved with any static allocation of 
$C$ distinct files. However, both LRU and LFU demonstrate equally disappointing performances. This is because at any time LRU stores the $C$
most recently requested files, while LFU retains the $C$ most frequently requested ones, but the next request is not for any of these cached files.

In contrast, both NFPL-Var and FPL showcase performances that are close to optimal. This reaffirms the resilience and adaptability of online learning policies across request processes as different as the three traces we considered. Intriguingly, NFPL-Var, which is inherently ``noisier,'' outperforms FPL to some extent. This phenomenon can be explained: the noisier  $\hat{\mathbf{r}}_{1:t} + \boldsymbol{\gamma}_t$, the more the cache tends to store a random set of files disregarding past requests. Such strategy is precisely up for the round-robin trace.

%In the round-robin trace, LRU and LFU behave poorly. This is because in this adversarial trace, at each time step, the requested file is neither one of the most $C$ popular items nor the $C$ most recently requested objects. On the other hand, NFPL-Var and FPL have a near-optimal performance with a slight advantage of NFPL-Var over FPL. This can be explained by observing that i) the policy RND, which selects the files to cache uniformly at random, is optimal in the round-robin trace and ii) the FPL algorithm bridges the gap between the policy RND and LFU and that as $f$ approaches $0$, NFPL-Var mimics the behavior of RND. 

% In all three sub-figures of Figure $1$, NFPL-Var outperforms classical policies (LRU and LFU). We also observe that the smaller is $f$, the smaller is the average miss ratio. Indeed, relying on recency or popularity can be misleading in this adversarial trace.
% Indeed, in this adversarial trace, the most recently requested files and the most popular files coincides and they are not going to be requested in the near future.

% In all three sub-figures of Figure $1$, NFPL-Var outperforms classical policies (LRU and LFU) in the round-robin trace. 
% Indeed, in this case, relying solely on popularity and recency is misleading. Meanwhile, using noisy popularity, as highlighted in Corollary~\ref{co:NFPL-Var}, leads to a near-optimal performance.

\subsection{NFPL-Fix vs. NFPL-Var}
\label{ss:NFPL-Fix-Vs-NFPL-Var}
We compare the performance of NFPL-Fix, NFPL-Var, and OPT on all the considered traces for two cache sizes: $C \in \{10, 200\}$ for the Zipf trace and $C \in \{10, 100\}$ for the Akamai and round-robin traces. Figure \ref{fig:NFPL-FIX-VAR} illustrates the average miss ratio for all the aforementioned caching policies when varying sampling probabilities, i.e., $f$ for NFPL-Var and $b/B$ for NFPL-Fix. 
%For each $f$, NFPL-Fix's parameter $b$ is set to $\lfloor B \cdot f \rfloor$. For each considered trace, we evaluate NFPL-Fix, NFPL-Var, and OPT 
  
%Across all considered traces, we observe that the distinction between NFPL-Fix and NFPL-Var in terms of the average miss ratio remains minimal across all the presented sampling probabilities. This suggests that the choice of the sampling method may have a negligible effect on NFPL performance.

Across the various traces we analyzed, the performance difference between NFPL-Fix and NFPL-Var is consistently minimal for all the sampling rates. This indicates that the selection of the sampling method may exert only a marginal impact on the performance of NFPL.

The influence of the sampling rate varies across the traces, aligning with the patterns previously noted in Figure~\ref{fig:hit-rates}. For the Zipf and Akamai traces, the performance of both NFPL-Fix and NFPL-Var tends towards that of OPT with increasing sampling rates. This is attributable to the relatively stationary nature of these traces, where the count of past requests serves as a good predictor for future requests; thus, more precise estimates bolster performance. In contrast, the round-robin trace benefits from noisier estimates, as it is preferable to overlook past requests in this scenario. As a result, the performance of NFPL-Fix and NFPL-Var deteriorates with a rising sampling rate.

%The effect of the sampling rate is different across the traces and consistent with what already observed in Figure~\ref{fig:hit-rates}. 
%Under the Zipf and Akamai traces, the performance of NFPL-Fix and NFPL-Var approach that of OPT as the sampling rate increases. This is because in relatively stationary traces, the number of past requests is the right metric to consider and more accurate estimates improve the performance. On the contrary, noisier estimates are more beneficial under the round-robin trace because past requests are better ignored. Correspondingly, NFPL-Fix and NFPL-Var performance worsen as the sampling rate increases.

%For example, when the sampling probability is $f = 0.15$ and the cache capacity is $C = 10$, the difference between OPT and NFPL-Fix is less than $0.05$ for the Zipf trace and less than $0.03$ for the Akamai trace. 
%On the other hand, NFPL-Fix and NFPL-Var seem to perform slightly better as $f$ decreases in the round-robin trace, which was already observed in Figure~\ref{fig:Average-Miss-Ratio-Round-Robin}, and explained by the fact that the caching policy RND, that selects uniformly at random files to cache, is optimal in this case, and that small values of $f$ helps  NFPL-Var to better mimic the behavior of RND. 

\section{Conclusion}\label{sec:conclusions}

In this paper, we introduce the Noisy-Follow-the-Perturbed-Leader (NFPL) algorithm, a variant of the Follow-the-Perturbed-Leader (FPL) algorithm that incorporates noisy cost estimates, and provide conditions on the cost estimates estimator for which NFPL achieves sublinear regret. In the context of the caching problem, we propose two NFPL algorithms, NFPL-Fix and NFPL-Var, based on sampling, that achieve sublinear regret. By conducting experiments on both synthetic and real world traces, we show the impact of request sampling on the performance of NFPL. In future work, we plan to investigate the regret of NFPL when the request estimator is based on approximate counting data structures such as the Count-Min Sketch \cite{mazziane2022analyzing}.

\bibliography{ref}
\bibliographystyle{ieeetr}
%S: I prefer to have the references ordered alphabetically

\end{document}